\documentclass[letterpaper, 10 pt, conference]{ieeeconf}  

\IEEEoverridecommandlockouts                              

\usepackage[nospace]{cite}
\usepackage{amsmath,amssymb,amsfonts}
\usepackage{graphicx}
\usepackage{textcomp}
\usepackage[dvipsnames]{xcolor}
\usepackage{multirow}
\usepackage{makecell}
\usepackage{tikz}
\usepackage{color, colortbl}
\usepackage{xcolor}
\usepackage{stfloats}
\usepackage{url}
\usepackage{caption}
\definecolor{LightBlue}{RGB}{212, 250, 252} 
\usepackage[flushleft]{threeparttable}
\usepackage{booktabs}
\usetikzlibrary{arrows.meta}
\let\labelindent\relax
\usepackage{enumitem}
\definecolor{LightBlue}{RGB}{212, 250, 252}
\usepackage{tabulary}
\usepackage{algorithm}
\usepackage{algpseudocode}
\usepackage{amsmath}

\def\mycircle[#1]{\tikz\draw[#1,fill=#1] (0,0) circle (0.125cm);}

\def\arrowright[#1]{\begin{tikzpicture}
  \draw[#1, -{Triangle[width = 5pt, length = 2pt]}, line width = 2pt] (0.0, 0.0) -- (0.25, 0.0);\path (current bounding box.south west) +(0,-0.06);
\end{tikzpicture}}

\def\arrowleft[#1]{\begin{tikzpicture}
  \draw[#1, -{Triangle[width = 5pt, length = 2pt]}, line width = 2pt] (0.25, 0.0) -- (0.0, 0.0);\path (current bounding box.south west) +(0,-0.06);

\end{tikzpicture}}

\def\arrowup[#1]{\begin{tikzpicture}
  \draw[#1, -{Triangle[width = 5pt, length = 2pt]}, line width = 2pt, rotate=270] (0.25, 0.0) -- (0.0, 0.0);
\end{tikzpicture}}

\def\arrowstraightleft[#1]{\begin{tikzpicture}
\draw[#1, -{Triangle[width = 5pt, length = 2pt]}, line width = 2pt, rotate=270] (0.25, 0.0) -- (0.0, 0.0);
  \draw[#1, -{Triangle[width = 5pt, length = 2pt]}, line width = 2pt] (0.0, -0.17) -- (-0.15, -0.17);
\end{tikzpicture}}

\def\arrowstraightright[#1]{\begin{tikzpicture}
    \draw[#1, -{Triangle[width = 5pt, length = 2pt]}, line width = 2pt, rotate=270] (0.25, 0.0) -- (0.0, 0.0);

  \draw[#1, -{Triangle[width = 5pt, length = 2pt]}, line width = 2pt] (0.0, -0.17) -- (0.15, -0.17);

\end{tikzpicture}}

\definecolor{Pistachio}{RGB}{140, 212, 126} 
\definecolor{CrayolaYellow}{RGB}{248, 214, 109}
\definecolor{PastelRed}{RGB}{255, 105, 97}
\definecolor{PastelOrange}{RGB}{255, 181, 76}
\definecolor{SoftCharcoal}{RGB}{66,66, 66}
\title{\LARGE \bf EffiComm: Bandwidth Efficient Multi Agent Communication}

\author{Melih Yazgan$^{1\star}$, Allen Xavier Arasan$^{3\star}$ and J. Marius Zöllner$^{1,2}$
\thanks{$^{1}$ FZI Research Center for Information Technology, Department of Technical Cognitive Systems, Haid-und-Neu Str. 10-14, Karlsruhe, Germany.
	{\tt\small \{surname\}@fzi.de}.}
\thanks{$^{2}$ Karlsruhe Institute of Technology (KIT), Institute of Applied Technical Cognitive Systems, Kaiserstr. 12, Karlsruhe Germany. {\tt\small \{prename.surname\}@kit.edu}.}
\thanks{$^{3}$ Karlsruhe Institute of Technology (KIT), Institute for Information Processing Technology, Engesserstr. 5, Karlsruhe Germany. {\tt\small \{prename.surname\}@student.kit.edu}.}
\thanks{$^\star$ Equal contribution}%
}

\begin{document}

\maketitle
\thispagestyle{empty}
\pagestyle{empty}

\begin{abstract}
Collaborative perception allows connected vehicles to exchange sensor information and overcome each vehicle’s blind spots. Yet transmitting raw point clouds or full feature maps overwhelms Vehicle-to-Vehicle (V2V) communications, causing latency and scalability problems. We introduce EffiComm, an end-to-end framework that transmits \(\mathbf{<40\%}\) of the data required by prior art while maintaining state-of-the-art 3-D object-detection accuracy. 
EffiComm operates on Bird’s-Eye-View (BEV) feature maps from any modality and applies a two-stage reduction pipeline: 
(1) Selective Transmission (ST) prunes low-utility regions with a confidence mask; 
(2) Adaptive Grid Reduction (AGR) uses a Graph Neural Network (GNN) to assign vehicle-specific keep ratios according to role and network load. 
The remaining features are fused with a soft-gated Mixture-of-Experts (MoE) attention layer, offering greater capacity and specialization for effective feature integration. 
On the OPV2V benchmark, EffiComm reaches 0.84-mAP@0.7 while sending only an average of $\sim1.5$ MB per frame, outperforming previous methods on the accuracy-per-bit curve. 
These results highlight the value of adaptive, learned communication for scalable Vehicle-to-Everything (V2X) perception.
\end{abstract}

\section{Introduction}
\label{sec:introduction} 

Recent progress in sensors, deep networks, and onboard computing power has brought reliable single-vehicle perception within reach~\cite{liang2018deep}.
However, adverse weather, occlusion and limited field-of-view still plague standalone systems~\cite{zhangAsync2023,yazgan2024real}.
Collaborative perception, in which vehicles and infrastructure nodes share observations, can fill those gaps by combining complementary viewpoints~\cite{yazgan2024real}.
The challenge lies in moving high-dimensional LiDAR or camera data across a bandwidth-constrained Vehicle-to-Everything (V2X) channel without flooding it~\cite{zhou2020evolutionary}.
Intermediate Bird’s-Eye-View (BEV) feature maps may exceed tens of megabytes per frame when naively broadcast.

We tackle this bottleneck with \textbf{EffiComm}, a modular framework that reduces transmitted bytes while keeping detection accuracy high.
EffiComm operates directly on BEV features, irrespective of the upstream backbone, and compresses them in two stages:

\begin{itemize}
    \item \textbf{Selective Transmission (ST):} a confidence mask retains only salient grid cells, discarding low-value regions before airtime is spent.
    \item \textbf{Adaptive Grid Reduction (AGR):} a Graph-Attention Network (GAT) predicts per-vehicle keep ratios conditioned on role, scene complexity and instantaneous network load.
\end{itemize}
The remaining features are fused on the ego vehicle with a soft-gated Mixture-of-Experts (MoE) attention block, designed to provide rich fusion capacity through specialized experts within a large parameter pool. Our contributions are as follows:
\begin{enumerate}[leftmargin=*]
    \item A bandwidth-aware, end-to-end pipeline that transmits \(\mathbf{<40\%}\) of the data required by prior works while preserving accuracy.
    \item A GNN-driven compression policy that assigns vehicle-specific keep ratios and adapts to dynamic traffic and link conditions.
    \item An MoE-attention fusion head that provides enlarged representational capacity, enabling effective fusion while maintaining high detection accuracy.
    \item Extensive evaluation on OPV2V dataset~\cite{xu2022opencood} showing the best accuracy-per-bit trade-off to date.
\end{enumerate}

The paper is organized as follows. Section~\ref{sec:related} reviews related work. Section~\ref{sec:methodology} details the two-stage compression and MoE fusion. Section~\ref{sec:evaluation} reports quantitative and ablation studies. Section~\ref{sec:future_works} discusses limitations and future extensions. Section~\ref{sec:conclusion} concludes.

\section{Related Work}\label{sec:related}
In this section, we discuss three important pillars in collaborative perception research:
i) the representation exchanged between agents,
ii) the policies for deciding when and how much information to transmit, and
iii) the mechanisms for fusing the received features into a coherent global understanding.

Accordingly, we structure our discussion into four sub-sections:
we first cover different levels of collaborative feature sharing,
then review communication-aware scheduling strategies,
next examine attention-based fusion mechanisms,
and finally highlight recent advances in MoE architectures that inform our design.
We identify the specific gap addressed by EffiComm.

\textbf{Feature-level collaboration.} Sharing raw LiDAR or camera data overwhelms V2X links, so recent work exchanges intermediate features.  
F-Cooper~\cite{chen2019fcooper} broadcasts every vehicle’s BEV tensor and stacks them, proving feasibility but still sending many background cells.  
DiscoGraph~\cite{li2021discograph} compresses features with 1-D convolutions yet keeps a fixed per-agent payload that may contain redundant regions.  
AttentiveFusion~\cite{xu2022opencood} adds an encoder–decoder for channel-wise compression, but it still transmits a spatially dense grid.  
Even with advanced channel reduction strategies utilized in methods like FFNet~\cite{yu2023ffnet}, Slim-FCP~\cite{guo2022slimfcp}, VIMI~\cite{wang2023vimi}, and ScalCompress~\cite{yuan2023scalcompress}, the full $H{\times}W$ lattice is still broadcast. 
Consequently, even state-of-the-art feature-level schemes continue to air substantial non-informative content.

\textbf{Communication-aware masking and scheduling.} Bandwidth scarcity in dense traffic prompted policies that adapt when or what to send.  
When2Com~\cite{liu2020when2com} learns a binary gate that decides whether any message is necessary; its successor Select2Col~\cite{liu2024select2col} leverages a spatial–temporal importance metric to cluster collaborators over time and skip frames whose marginal value is low.  
These methods act in the time dimension but still ship the entire grid once activated. Where2Comm~\cite{hu2022where2comm} moves the decision into the spatial domain, masking low-confidence BEV cells and roughly halving bytes on OPV2V.  
BM2CP~\cite{zhao2023bm2cp} applies a similar idea to camera–LiDAR fusion.  
Spatial masking is effective, but current variants use identical thresholds for every car and frame; when many vehicles contend for the channel, overall throughput can still spike because keep ratios are chosen locally and independently.

\textbf{Attention-based fusion.}
Transformer-style attention for V2X fusion has become the de facto architecture for aligning multi-agent features once they are received.  
AttentiveFusion~\cite{xu2022opencood} inserts multi-head self-attention between stacked BEV maps and shows robustness to pose error and latency.  
V2X-ViT~\cite{xu2022v2xvit} alternates inter-agent and spatial attention in a Vision-Transformer backbone; it still expects spatially dense maps from every peer.
Liu et al.~\cite{liu2024select2col} learn agent-level importance weights, whereas EdgeCooper~\cite{ye2023edgecooper} models vehicles as nodes in a GAT.  
These transformers excel at information integration but leave the upstream bandwidth bottleneck unsolved: they optimise accuracy rather than transmission efficiency. MoE layers expand capacity through conditional routing with modest compute cost~\cite{shazeer2017outrageously,riquelme2021scaling}.  
MoE is mainstream in NLP and single-image vision, yet remains largely unexplored in bandwidth-constrained multi-agent perception.  
Existing cooperative frameworks employ dense attention and do not exploit expert specialisation to encourage upstream sparsity.
\section{Methodology}
\label{sec:methodology}
\begin{figure*}[ht]
    \centering
    \includegraphics[width=0.8\linewidth]{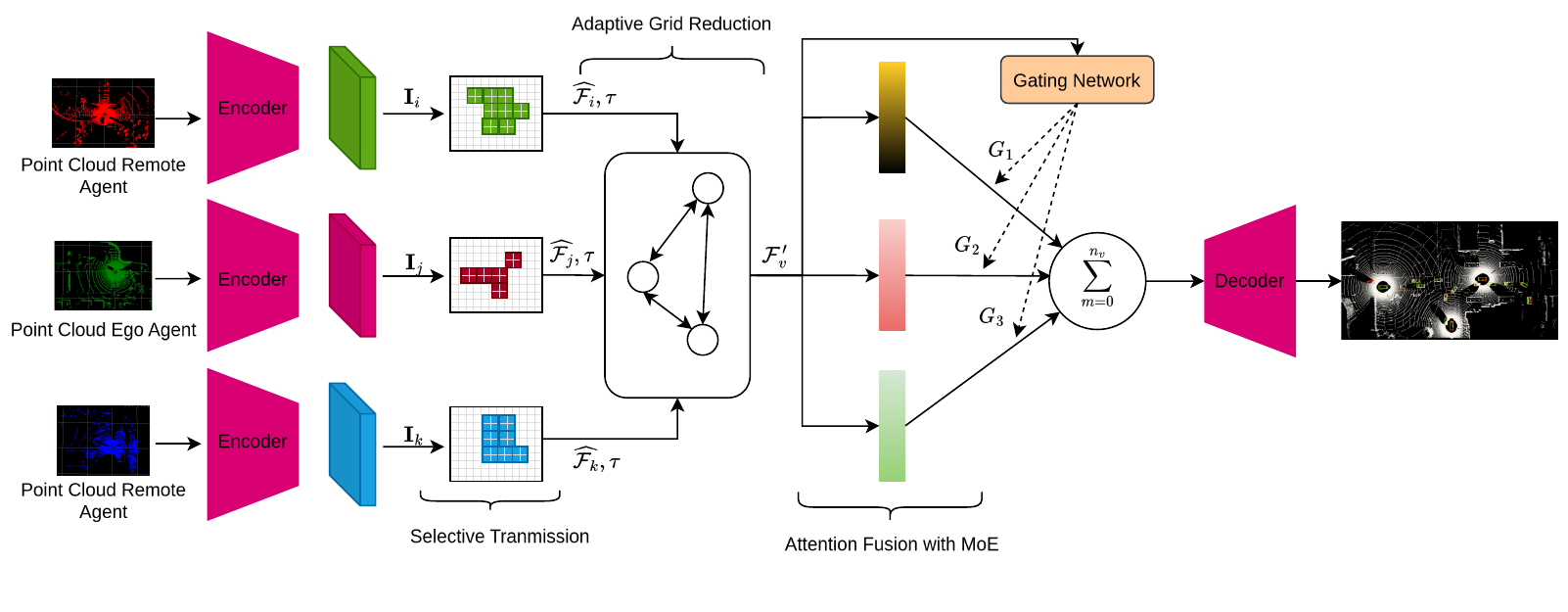}
    \caption{End-to-end pipeline illustrating feature extraction, the two-stage reduction process (Selective Transmission and Adaptive Grid Reduction), MoE-based Feature Fusion, and final Detection output for multiple collaborating vehicles.}
    \label{fig:efficomm}
\end{figure*}

Figure \ref{fig:efficomm} shows the EffiComm pipeline, with Algorithm 1 listing each step in sequence. EffiComm addresses the tight V2X bandwidth budget by reducing the size of the BEV feature maps that vehicles have to share. It first drops low-confidence cells via ST, then applies a graph-based, context-aware keep-ratio predictor, AGR, to trim each remote map even further. Together, the two stages remove redundant or low-utility information, cut the transmitted payload to a fraction of its original size, and still keep the spatial cues needed for accurate 3-D object detection.
\subsection {Feature Encoder}

\paragraph{LiDAR preprocessing}
All methods compared in Section~\ref{sec:evaluation} use the same LiDAR-based PointPillars backbone.  
To ensure that any performance gap comes solely from the
communication scheme, and not from a stronger or weaker perception
backbones, we instantiate EffiComm with the same
PointPillars network of~\cite{lang2019pointpillars}.
LiDAR point clouds  are first partitioned into
vertical pillars of fixed ground-plane resolution
\(\Delta_x\times\Delta_y\).  A PointNet‐style MLP encodes each pillar
and max-pooled along the height axis, producing a dense 2-D pseudo-image that
can be processed by ordinary CNN layers.  We adopt the four-stage
PointPillar backbone proposed in work~\cite{lang2019pointpillars} without any
architectural modifications.

\paragraph{BEV feature tensor}
For every time step and every vehicle the backbone yields a Bird’s-Eye-View
(BEV) tensor \(\mathcal{F}\;\in\;\mathbb{R}^{L\times H\times W}\), where \(L\) is the number of channels and \(H\times W\) the spatial grid
size.  Because each pixel \((i,j)\) in \(\mathcal{F}\) corresponds to a fixed
cell in world coordinates, later masking operations can zero individual
locations without breaking geometric consistency.

\paragraph{Class-confidence map}
The detection head returns per-class objectness logits in the same layout; we denote them
$\mathcal{C}\in\mathbb{R}^{L\times H\times W}$.
Taking the channel-wise maximum collapses $\mathcal{C}$ into a
single-channel importance image
\[
  \mathbf{I}\;=\;\max_{\,\ell}\,\mathcal{C}_{\ell}\in\mathbb{R}^{1\times H\times W}
\]
which is the sole side input required by the two bandwidth-reduction stages below.
\subsection{Selective Transmission}
\label{sec:selective_transmission}
Selective Transmission is the first bandwidth gate in EffiComm.
Starting from the importance image $\mathbf{I}$ supplied by the
feature encoder, it removes spatial cells that are unlikely to benefit
downstream detection.

\paragraph{Binary mask}
After sigmoid normalisation, each pixel is compared with a global threshold
$\mu$ (as in~\cite{hu2022where2comm}):
\begin{equation}
  M[i,j] \;=\;
  \begin{cases}
    1 & \text{if }\sigma\!\bigl(\mathbf{I}[i,j]\bigr) > \mu,\\[2pt]
    0 & \text{otherwise},
  \end{cases}
  \label{eq:mask1}
\end{equation}
yielding a mask $M\in\{0,1\}^{1\times H\times W}$ that is broadcast over
channels and multiplied with the feature tensor in, $\widehat{\mathcal{F}}=\mathcal{F}\odot M$. The ego map is never transmitted, therefore we set $M_{\text{ego}}=1$.

It is important to note the different strategies employed during training and inference for this module.
Training uses a random top-K selection (where the number of kept cells $K$ is sampled uniformly, $K \sim \mathrm{Uniform}(0,H \times W)$ based on the confidence map as a form of data augmentation and regularization.
Inference, however, (for both validation and deployment) deterministically applies the fixed global threshold $\mu$ as defined in Eq.~\eqref{eq:mask1}, ensuring only cells with confidence $\sigma(\mathbf{I}[i,j])>\mu$ are kept.
This approach ensures the network learns robustness to varying levels of input sparsity during training, while evaluation remains consistent and deterministic based on the chosen threshold $\mu$.

\paragraph{Instantaneous transmission rate}
We define the instantaneous transmission rate~$\tau$ as the fraction of grid cells retained by the inference mask  $M$
\begin{equation}
  \tau \;=\;
  \frac{\sum_{i,j}M[i,j]}{H\,W}
  \label{eq:tr}
\end{equation}
acts as a coarse measure of link load.
The pair $(\widehat{\mathcal{F}},\tau)$ is forwarded to
\textbf{AGR},
Section~\ref{sec:adaptive_grid}, which refines sparsity per vehicle
based on graph attention and the very same rate~$\tau$.

\subsection {Adaptive Grid Reduction}
\label{sec:adaptive_grid}
\begin{figure*}[htbp]
    \centering
    \includegraphics[width=0.8\linewidth]{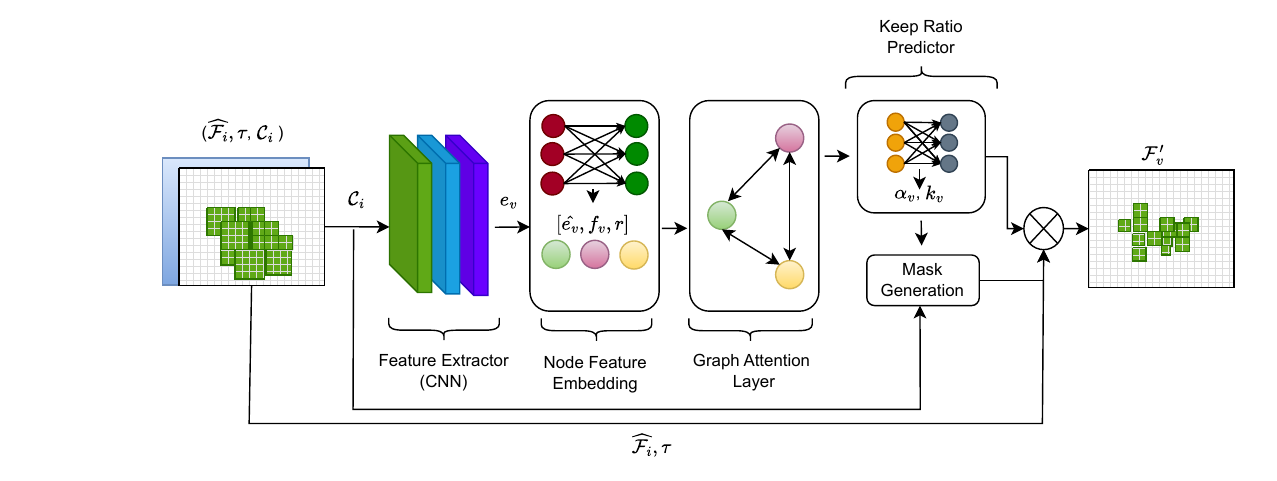}
    \caption{Internal components of the adaptive reduction stage. A GNN processes vehicle features (derived from confidence maps) and network context to dynamically calculate a keep ratio ($k_v$), which determines the mask applied to produce the reduced feature map $\mathcal{F}'_v$.} 
    \label{fig:adp}
\end{figure*}

While ST provides a general mechanism for reducing feature density, the AGR module, shown in Fig. \ref{fig:adp}, adds context-awareness by dynamically adjusting compression levels based on vehicle-specific characteristics and network conditions. This module, outlined in Algorithm 2, recognizes that different vehicles and scenarios may require different levels of feature preservation.

The core of this module is a GNN that models relationships between vehicles and determines appropriate keep ratios. For each vehicle, representative features from their confidence maps using a CNN are extracted, which is given as $e_v = \phi(\mathcal C_v)$. These high-dimensional features are embedded into a lower-dimensional space using a Multi Layer Perceptron (MLP) represented as $\hat{e}_v = \psi(e_v)$.
To incorporate contextual information beyond just the feature content, node features including vehicle type and current transmission rate are constructed using $n_v = [\hat{e}_v, f_v, \tau]$ where $f_v$ is a binary flag and $r$ represents the overall transmission rate calculated from the ST module. These node features are processed by a GAT that enables message passing between different vehicles, effectively allowing them to coordinate their compression strategies given by Eq. {\ref{eq:gat}} where  $\mathbf{n_v}$ represent the nodes features, where every node represents a vehicle in the scene. $\mathbf{A}$ is the adjacency matrix defining the connectivity between vehicles in a fully-connected communication graph.
Traditional perception models often attempt to process comprehensive global information, which leads to the inclusion of irrelevant or redundant information \cite{song2025insightdriveinsightscenerepresentation}. Hence, we establish a base keep ratio to ensure critical information is preserved while filtering out redundant features. For the ego vehicle, we set this ratio at $k_\text{{ego}} = 0.9$ to maintain high fidelity of its immediate perceptual field, which is essential for safety-critical decision-making. 
\begin{equation}
\mathbf{z_v} = \text{GAT}(\mathbf{n_v}, \mathbf{A})
\label{eq:gat}
\end{equation}
\begin{minipage}{\columnwidth}
\hrule height 1pt
\vskip 4pt
\textbf{Algorithm 1: EffiComm for Bandwidth-Efficient Communication}
\vskip 2pt
\footnotesize
\begin{tabular}{r@{\quad}l}
\label{algo:efficomm}\\
1. & \textbf{Input:}                                                        \\[-1pt]
2. & \qquad$\mathcal{F}$  – BEV feature tensors of all vehicles \\[-1pt]
3. & \qquad$\mathcal C$  – class-confidence tensors (same layout) \\[-1pt]
4. & \qquad\emph{hyper-parameters} for the reduction modules \\[2pt]
5. & \textbf{Output:} $\mathcal F_{\text{out}}$ – fused, bandwidth-reduced BEV\\[4pt]

6. & \textbf{Procedure} \textsc{EffiComm}$(\mathcal F,\mathcal C,\text{params})$      \\
7. & \qquad\textbf{for each} batch $b$ \textbf{do}                                  \\
8. & \qquad\quad$\mathcal M_b,\,\tau_b \leftarrow$ \textsc{GenerateTransmissionMask}$(\mathcal C_b)$\\[-1pt]
9. & \qquad\quad\textit{/* first-stage spatial pruning */}                         \\
10.& \qquad\quad$\mathcal F_b \leftarrow \mathcal F_b \odot \mathcal M_b$            \\

11.& \qquad\quad\textit{/* second-stage adaptive thinning */}                       \\
12.& \qquad\quad$\mathcal F_b \leftarrow$ \textsc{ReduceFeatures}($\mathcal{F}_b$, $\tau_b$) \\[3pt]

13.& \qquad\quad$\mathcal V_b \leftarrow$ \textsc{GroupByVehicle}($\mathcal{F}_b$)     \\

14.& \qquad\quad$\mathcal F_b \leftarrow$ \textsc{FuseFeatures}$(\mathcal V_b)$       \\[2pt]
15.& \qquad\quad\textit{/* store fused map for this batch */}                        \\[-1pt]
16.& \qquad\quad $\mathcal F_b$ to $\mathcal F_{\text{out}}$                   \\[1pt]
17.& \qquad\textbf{end for}                                                          \\[2pt]
18.& \qquad\Return $\mathcal F_{\text{out}}$                                         \\[3pt]
19.& \textbf{End Procedure}                                                          \\[5pt]

20.& \textbf{Function} \textsc{GenerateTransmissionMask}$(\mathcal C)$               \\[-1pt]
21.& \qquad importance scores from $\mathcal C$                               \\[-1pt]
22.& \qquad / top-$K$ selection $\rightarrow$ binary mask $\mathcal M$      \\[-1pt]
23.& \qquad transmission rate $\tau$                                          \\[-1pt]
24.& \qquad\Return $(\mathcal M,\tau)$                                               \\[-1pt]
25.& \textbf{End Function}                                                           \\[5pt]

26.& \textbf{Function} \textsc{ReduceFeatures}$(\mathcal{F},\tau)$                    \\[-1pt]
27.& \qquad\textbf{for each} vehicle $v$ \textbf{do}                                 \\[-1pt]
28.& \qquad node feature $\mathbf n_v$ (CNN + MLP)                        \\[-1pt]
29.& \qquad\quad$\mathbf z_v \leftarrow \text{GAT}(\mathbf n_v)$                     \\[-1pt]
30.& \qquad keep ratio $k_v$ from $\mathbf z_v$ and~$\tau$               \\[-1pt]
31.& \qquad binary mask that preserves $k_v$ of highest-score cells        \\[-1pt]
32.& \qquad\textbf{end for}                                                          \\[-1pt]
33.& \qquad\Return bandwidth-reduced $\mathcal{F}$                                    \\[-1pt]
34.& \textbf{End Function}                                                           \\[5pt]

35.& \textbf{Function} \textsc{FuseFeatures}$(\mathcal V)$                           \\[-1pt]
36.& \qquad MoE-based scaled-dot attention across vehicles                            \\[-1pt]
37.& \qquad\Return fused map                                                        \\[-1pt]
38.& \textbf{End Function}                                                           \\
\end{tabular}
\vskip 4pt
\hrule height 1pt
\end{minipage}
\vspace{0.1cm}

For remote vehicles, we employ a more aggressive $k_\text{{remote}} = 0.5$ ratio, as remote vehicles primarily contribute supplementary context to the collaborative perception framework rather than direct decision inputs, allowing for greater compression without sacrificing overall system performance. The base keep ratios are modified according to vehicle type and network conditions. The final keep ratio is computed as given in Eq. {\ref{eq:kp}} where a keep ratio adjustment factor, determined from the GAT output, is given by $\alpha_v = \sigma(\text{MLP}(z_v))$.

\footnotesize
\begin{equation}
k_v = 
\begin{cases}
k_{ego} \times (0.5 + 0.5\alpha_v) \times (0.7 + 0.3(1-\tau)), & \text{if } v \text{ is ego} \\
k_{remote} \times (0.5 + 0.5\alpha_v) \times (0.7 + 0.3(1-\tau)), & \text{otherwise}
\end{cases}
\label{eq:kp}
\end{equation}
\normalsize
 The term $(0.5 + 0.5\alpha_v)$ applies the GAT-derived adjustment, allowing the keep ratio to be scaled between 50\% and 100\% of its base value depending on feature importance. The term $(0.7 + 0.3(1-\tau))$ incorporates transmission rate awareness by reducing the keep ratio when network congestion increases. Rate-distortion \cite{1091311} theory establishes that when approaching channel capacity limits, optimal compression requires increasing distortion to maintain feasible transmission rates. Our feedback mechanism Eq. \ref{eq:kp} implements this principle by reducing keep ratios proportionally to instantaneous transmission rate $\tau$, ensuring that higher congestion triggers more aggressive compression while preserving critical detection information.
 For each vehicle, the number of grid cells to keep is calculated using Eq. {\ref{eq:grid}}.
\begin{equation}
K_v = \max(1, \lfloor H \times W \times k_v \rfloor)
\label{eq:grid}
\end{equation}
A vehicle specific mask $M_v$ is then created by selecting the $K_v$ grid cells with the highest importance scores in the confidence map $\mathcal{C}_v$. This mask is applied to the feature map $\hat{\mathcal{F}}_v$ (output of Selective Transmission) as shown in Eq.~\eqref{eq:rd}. 
\begin{equation}
\mathcal{F}'_v = \hat{\mathcal{F}}_v \odot M_v 
\label{eq:rd}
\end{equation}

Note that this adaptive mask $M_v$, determined by the keep-ratio $k_v$ and confidence map $\mathcal{C}_v$, is applied to the feature map $\hat{\mathcal{F}}_v$ which may have already been sparsified by the preceding Selective Transmission stage (Sec.~\ref{sec:selective_transmission}).
The adaptive nature of this approach allows EffiComm to maintain an effective balance between perception quality and bandwidth usage across diverse driving scenarios and varying network condition. This reduced feature $\mathcal{F}'_v$ is then transmitted from the remote vehicle to the ego vehicle, followed by feature fusion in the ego vehicle. 
Qualitative pruning results are compared against Where2Comm in Fig.~\ref{fig:WvsE} and discussed in Sec.~\ref{sec:evaluation}.


\vspace{0.2cm}
\begin{minipage}{0.95\columnwidth}
 \hrule height 1pt
 \vskip 4pt
  \textbf{Algorithm 2: Adaptive Grid Reduction}
  \vskip 4pt
  \footnotesize
  \begin{tabular}{r@{\quad}l}
  \label{algo:adp}
  1.  & \textbf{Input:} \\
  2.  & \quad $\mathcal{F} = \{\mathcal{F}_{1},\mathcal{F}_{2},...,\mathcal{F}_{N}\}$: Feature maps from $N$ vehicles \\ 
  3.  & \quad $\mathcal{C} = \{C_1, C_2, \ldots, C_N\}$: Confidence maps \\
  4.  & \quad $k_{ego}$: Base keep ratio for ego vehicle \\
  5.  & \quad $k_{remote}$: Base keep ratio for remote vehicles \\
  6.  & \quad $\tau$): Network transmission rate \\ 
  7.  & \textbf{Output:} \\
  8.  & \quad $\mathcal{F}' = \{\mathcal{F}'_1, \mathcal{F}'_2, \ldots, \mathcal{F}'_N\}$: Reduced feature maps \\ 
  9.  & \textbf{Function:} \textsc{AdaptiveGridReduction}($\mathcal{F}, \mathcal{C}, k_{ego}, k_{remote}, \tau$) \\ 
  10. & \quad \textbf{for} each vehicle $v \in \{1,2,\ldots,N\}$ \textbf{do} \\
  11. & \quad\quad $e_v \gets \phi(C_v)$ \quad \Comment{Extract features via CNN} \\
  12. & \quad\quad $\hat{e}_v \gets \psi(e_v)$ \quad \Comment{Embed into lower dimension} \\
  13. & \quad \textbf{end for} \\
  14. & \quad Define vehicle type flags: $f_v \gets \begin{cases} 1, & \text{if } v \text{ is ego} \\ 0, & \text{otherwise} \end{cases}$ \\
  15. & \quad Construct node features: $n_v \gets [\hat{e}_v, f_v, \tau]$ \\
  16. & \quad Build adjacency matrix $\mathbf{A}$ for vehicle graph \\
  17. & \quad Apply graph convolution: $\mathbf{z_v} \gets \text{GAT}(\mathbf{n_v}, \mathbf{A})$ \\
  18. & \quad \textbf{for} each vehicle $v \in \{1,2,\ldots,N\}$ \textbf{do} \\
  19. & \quad\quad $\alpha_v \gets \sigma(\text{MLP}(z_v))$ \quad \Comment{Adjustment factor} \\
  20. & \quad\quad Calculate keep ratio $k_v$ using Equation~\ref{eq:kp}\\
  21. & \quad\quad Keep ratio: $k_v \gets \max(\min(k_v, 0.95), 0.1)$ \\
  22. & \quad\quad Grid points to preserve: $K_v \gets \max(1, \lfloor H \times W \times k_v \rfloor)$ \\
  23. & \quad\quad Importance mask $M_v$ by selecting top $K_v$ points from $C_v$ \\
  24. & \quad\quad Apply mask to features: $\mathcal{F}'_v \gets \hat{\mathcal{F}}_v \odot M_v$ \\ 
  25. & \quad \textbf{end for} \\
  26. & \quad \Return $\mathcal{F}' = \{\mathcal{F}'_1, \mathcal{F}'_2, \ldots, \mathcal{F}'_N\}$ \\ 
  27. & \textbf{End Function} \\
  \end{tabular}
  \vskip 5pt
  \hrule height 1pt
\end{minipage}
\vspace{0.1cm}
\subsection {Feature Fusion}
Once features are received (potentially after reduction by the previous stages), this module addresses the challenge of integrating heterogeneous information from different viewpoints using a MoE architecture. The MoE approach allows for specialized processing by employing multiple parallel expert networks alongside a gating network. The input query $q$, key $k$, and value $v$ for the attention computations are derived from the received feature maps $\mathcal{F}'_{v}$. 

Our MoE architecture introduces specialized expert networks, where each expert is implemented as a standard Scaled Dot-Product Attention module. The fusion process is governed by Equation~\eqref{eq:moe_fusion}: 

\begin{equation}
\label{eq:moe_fusion} 
\text{$X$} = \sum_{i=1}^{E} G(\mathcal{F}'_{v})_{i} \cdot W_{i}(\mathcal{F}'_{v}) 
\end{equation}

Here, $W_{i}(\mathcal{F}'_{v})$ represents the output of the $i$-th Scaled Dot-Product Attention expert, and $G(\mathcal{F}'_{v})_{i}$ is the corresponding gating weight assigned by a dedicated gating network. 
The gating network first computes logits $L(x)$ for each expert based on the input features $x$ (derived from $\mathcal{F}'_{v}$) using a learned MLP: 

\begin{equation}
\label{eq:gating_logits} 
L(x) = \text{RouterMLP}(x) 
\end{equation}

The final gating weight for the $i$-th expert, $G(x)_i$, is then obtained by applying the Softmax function to these logits:

\begin{equation}
\label{eq:gating_softmax} 
G(x)_i = \text{softmax}(L(x))_i = \frac{\exp(L_i(x))}{\sum_{j=1}^{E} \exp(L_j(x))}
\end{equation}

The gating network dynamically soft routes information based on contextual relevance by determining these weights from a pooled representation of the input features. This allows the network to assess the context for each spatial location and assign appropriate weights, creating natural expert specialization during training. The individual Scaled Dot-Product Attention experts can thus focus on different attributes or patterns within the features. This MoE strategy enables vehicles to engage in more nuanced feature aggregation, allowing distinct information aspects to receive appropriate expert attention and enhancing the system's robustness in collaborative perception scenarios.

\textbf{Multi-Scale Fusion Strategy:} Our implementation leverages feature maps extracted by the backbone network at multiple resolutions. The sparsity pattern determined by the two-stage reduction pipeline is applied, and the subsequent MoE-based attention fusion described above occurs independently at each relevant feature scale $l \in \{1, ..., L\}$. 
The resulting fused feature maps from each scale are then combined, typically involving upsampling via deconvolution layers followed by concatenation, to form a final, rich feature representation. This representation is then passed to the detection decoder. This multi-scale fusion approach allows the system to integrate information captured at different levels of detail and abstraction.
\subsection{Feature Decoder}
The feature decoder transforms fused features $X$ into objects with class and regression outputs for perception results. For ego $v$, detections are obtained as $O^{(v)} = \text{dec}(X^{(v)})$, where each detection is a 7-dimensional vector $(c,x,y,h,w,\cos\theta,\sin\theta)$ representing class confidence, position, size, and angle cohesive with \cite{hu2022where2comm}.

\section{Experiments}
\label{sec:evaluation}

We evaluate EffiComm on the OPV2V dataset~\cite{xu2022opencood} and assess generalization on the Culver City dataset (also from~\cite{xu2022opencood}), which presents real-world driving scenarios. OPV2V includes diverse urban and suburban environments generated using OpenCDA~\cite{xu2021opencda} and CARLA~\cite{carla}, enabling controlled studies across varying traffic, weather, and agent conditions. It supports detailed analyses of multi-agent collaboration and perception under different policies. A recent survey by Yazgan et al.~\cite{YazganCoPeData} provides a broader overview of cooperative perception datasets.

EffiComm employs an end-to-end training strategy tailored for cooperative perception. Gradients are backpropagated through the entire network, including the PointPillars backbone, the selective transmission module, adaptive grid reduction, and the MoE fusion head. This eliminates the need for separate pre-training, improving training efficiency.

Experiments were conducted on two NVIDIA RTX 3090 GPUs using the Adam optimizer with an initial learning rate of 0.001 and batch size of 4. A cosine annealing scheduler with warm-up was employed, starting from $2\times10^{-5}$ for 10 epochs and decaying to $5\times10^{-6}$, balancing performance and computational cost. During inference, the ST module (Sec.~\ref{sec:selective_transmission}) utilizes a fixed confidence threshold $\mu = 0.01$. Applying this threshold to the sigmoid-normalized confidence map ($\sigma(I)$) primarily filters out grid cells where the model assigns extremely low confidence, acting as a conservative initial pruning step before the AGR stage.  For the primary results presented in this work (Tables~\ref{tab:results_method_comparison}-\ref{tab:effi_fusion_comparison}), the network was trained end-to-end using a detection loss function ($L_{det}$) consistent with the underlying PointPillars architecture~\cite{lang2019pointpillars}. Specific investigations into the effects of additional loss terms are detailed in Section~\ref{sec:loss_impact}. 

\subsection{Evaluation Metrics}
To assess communication efficiency, we quantify transmission volume by extracting non-zero elements from the feature maps ($\mathcal{F}_v$) generated by the backbone before fusion/communication. Following established practices \cite{zhao2023bm2cp, hu2022where2comm}, we compute the $\log_2$ of this count to represent the communication cost ('Comm') in our primary comparison table, simulating realistic data compression effects. Detection performance is evaluated using Average Precision (AP) at IoU thresholds of 0.5 and 0.7 on the OPV2V dataset. Additionally, we directly measure the raw bandwidth consumption (mean, standard deviation, max, min) in Megabytes (MB) per frame across the test set for a more detailed efficiency analysis. We utilized open-source weights for the AttentiveFusion and FCooper implementations, whereas the Where2Comm model was trained independently.

\subsection{Visual comparison with \textit{Where2Comm}}
\label{sec:qualitative_pruning}

\begin{figure}[!ht]
  \centering
  \begin{minipage}{0.48\textwidth}
    \centering
    \includegraphics[width=\linewidth]{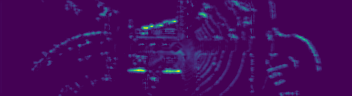}\\[-0.3em]
    \footnotesize\textbf{(a)} Where2Comm
  \end{minipage}

  \vspace{0.5em}

  \begin{minipage}{0.48\textwidth}
    \centering
    \includegraphics[width=\linewidth]{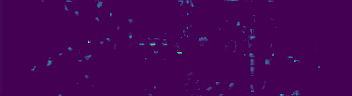}\\[-0.3em]
    \footnotesize\textbf{(b)} EffiComm
  \end{minipage}

  \caption{Comparison of feature selection between Where2comm and EffiComm. EffiComm selects fewer but more relevant features from other agents to the ego vehicle.}
  \label{fig:WvsE}
\end{figure}

To illustrate how the two-stage policy of EffiComm affects the spatial
footprint that is actually broadcast, Fig.~\ref{fig:WvsE}
(\textit{a}) shows the selected features produced by
Where2Comm~\cite{hu2022where2comm}, and
Fig.~\ref{fig:WvsE} (\textit{b}) the masked features obtained after both
selective transmission and adaptive grid reduction.
    
    
\subsection{Performance Comparison}
As demonstrated in Table~\ref{tab:results_method_comparison}, our proposed EffiComm approach achieves better detection performance (IoU@0.5: 0.92, IoU@0.7: 0.843) than state-of-the-art methods like AttentiveFusion and Where2Comm. Also, EffiComm significantly reduces the communication volume ('Comm') to \textbf{10.88}, outperforming AttentiveFusion (13.5) , FCooper (13.22), and Where2Comm (12.25). This highlights EffiComm's superior communication efficiency and detection in collaborative perception tasks based on the $\log_2$ metric.

\begin{table}[htbp]
\centering
\begin{tabular}{l|ccc} 
\toprule
\textbf{Method} & \textbf{AP@0.5} & \textbf{AP@0.7} & \textbf{Comm ($\log_2$ count)} \\
\midrule
AttentiveFusion  & 0.90 & 0.81 & 13.5 \\
FCooper & 0.88 & 0.79 & 13.22\\
Where2comm & 0.91 & 0.83 & 12.25 \\
EffiComm(Ours) & \textbf{0.92} & \textbf{0.843} & \textbf{10.89} \\
\bottomrule
\end{tabular}
\caption{Experimental Results on OPV2V test set}
\label{tab:results_method_comparison}
\end{table}
To evaluate EffiComm's generalization ability, we present results on the Culver City dataset. Table~\ref{tab:culver_city_results} shows the performance comparison with baseline methods. EffiComm achieves competitive detection performance on the Culver City dataset while demonstrating significantly lower communication cost compared to the baseline methods.

\begin{table}[htbp]
    \centering
    \begin{tabular}{l|ccc}
        \toprule
        \textbf{Method} & \textbf{AP@0.5} & \textbf{AP@0.7} & \textbf{Comm ($\log_2$ count)} \\
        \midrule
        AttentiveFusion & 0.85 & 0.73 & 13.19 \\
        FCooper & 0.84 & 0.73 & 12.91 \\
        Where2Comm & \textbf{0.87} & \textbf{0.75} & 12.42 \\
        EffiComm (Ours) & 0.86 & 0.74 & \textbf{10.99} \\
        \bottomrule
    \end{tabular}
\caption{Performance comparison on Culver City dataset}
\label{tab:culver_city_results}
\end{table}

\subsection{Bandwidth Efficiency Analysis}
A deeper analysis using direct bandwidth measurements further underscores EffiComm's efficiency advantage, as demonstrated in Table~\ref{tab:fusion_comparison}. EffiComm demonstrates the lowest average per-frame communication cost at merely \textbf{1.90 MB}. This represents a substantial improvement compared to AttentiveFusion (11.64) and Where2Comm (4.9 MB), transmitting approximately 2.5 times less data per frame than the latter.

Furthermore, EffiComm's bandwidth utilization is highly consistent, evidenced by its minimal standard deviation of \textbf{0.19 MB}, compared to 3.94 MB for AttentiveFusion and 0.35 MB for Where2Comm. The maximum observed bandwidth for EffiComm (2.66 MB) is less than half of Where2Comm’s peak (6.09 MB) and significantly lower than other baselines. EffiComm also achieves the lowest minimum per-frame bandwidth (1.65 MB). These metrics collectively demonstrate EffiComm's ability to maintain consistently low and stable communication loads, effectively mitigating bandwidth spikes.

\begin{table}[htbp]
\centering
\footnotesize
\begin{tabular}{l|cccc}
\toprule
\textbf{Method} & \textbf{Mean} & \textbf{Std} & \textbf{Max} & \textbf{Min}  \\
\midrule
AttentiveFusion  & 11.64 &3.94   & 22.8 & 4.52 \\
FCooper          & 9.588 & 3.203 & 18.8 & 3.6  \\
Where2comm       & 4.9   & 0.37  & 6.09 & 4.3 \\
EffiComm(Ours)   & \textbf{1.90} & \textbf{0.19} & \textbf{2.44} & \textbf{1.65} \\
\bottomrule
\end{tabular}
\caption{Feature bandwidth statistics in MB on OPV2V test dataset}
\label{tab:fusion_comparison}
\end{table}

\subsection{Impact of Mixture of Experts Fusion}
\label{sec:moe_impact} 
To isolate the effect of the fusion mechanism, we compared our proposed MoE module against standard Scaled Dot-Product Attention (SDPA) fusion ('EffiComm without MoE'). Both variants used the same reduction pipeline and were trained identically optimizing only the detection loss, ensuring a fair comparison of the fusion heads. Results are presented in Table~\ref{tab:effi_fusion_comparison}. While both methods achieve comparable detection accuracy (0.843 AP@0.7 for MoE vs. 0.838 for SDPA), the MoE variant exhibits significantly better communication efficiency, reducing the mean bandwidth from 2.33 MB/frame to 1.90 MB/frame and showing lower variance.

\begin{table*}[ht]
\centering
\footnotesize
\begin{tabular}{l|c c c c}
\toprule
\textbf{EffiComm Variant (Detection Loss Only)} & \textbf{AP@0.7} & \textbf{Comm ($\log_2$ count)} & \textbf{Mean (MB)} & \textbf{Std (MB)} \\
\midrule
EffiComm without MoE (Std. SDPA) & 0.838 & 11.18 & 2.33 & 0.27 \\ 
EffiComm (with MoE) & \textbf{0.843} & \textbf{10.89} & \textbf{1.90} & \textbf{0.18} \\ 
\bottomrule
\end{tabular}
\caption{EffiComm fusion method comparison on OPV2V test set (Models trained with detection loss only). Performance measured by AP@0.7. Communication cost ('Comm') reported as $\log_2(\text{non-zero elements} \times B)$. Bandwidth statistics Mean and Std reported in MB/frame. SDPA refers to Scaled Dot-Product Attention.}
\label{tab:effi_fusion_comparison} 
\end{table*}

\begin{figure}
    \centering
    \includegraphics[width=1\linewidth]{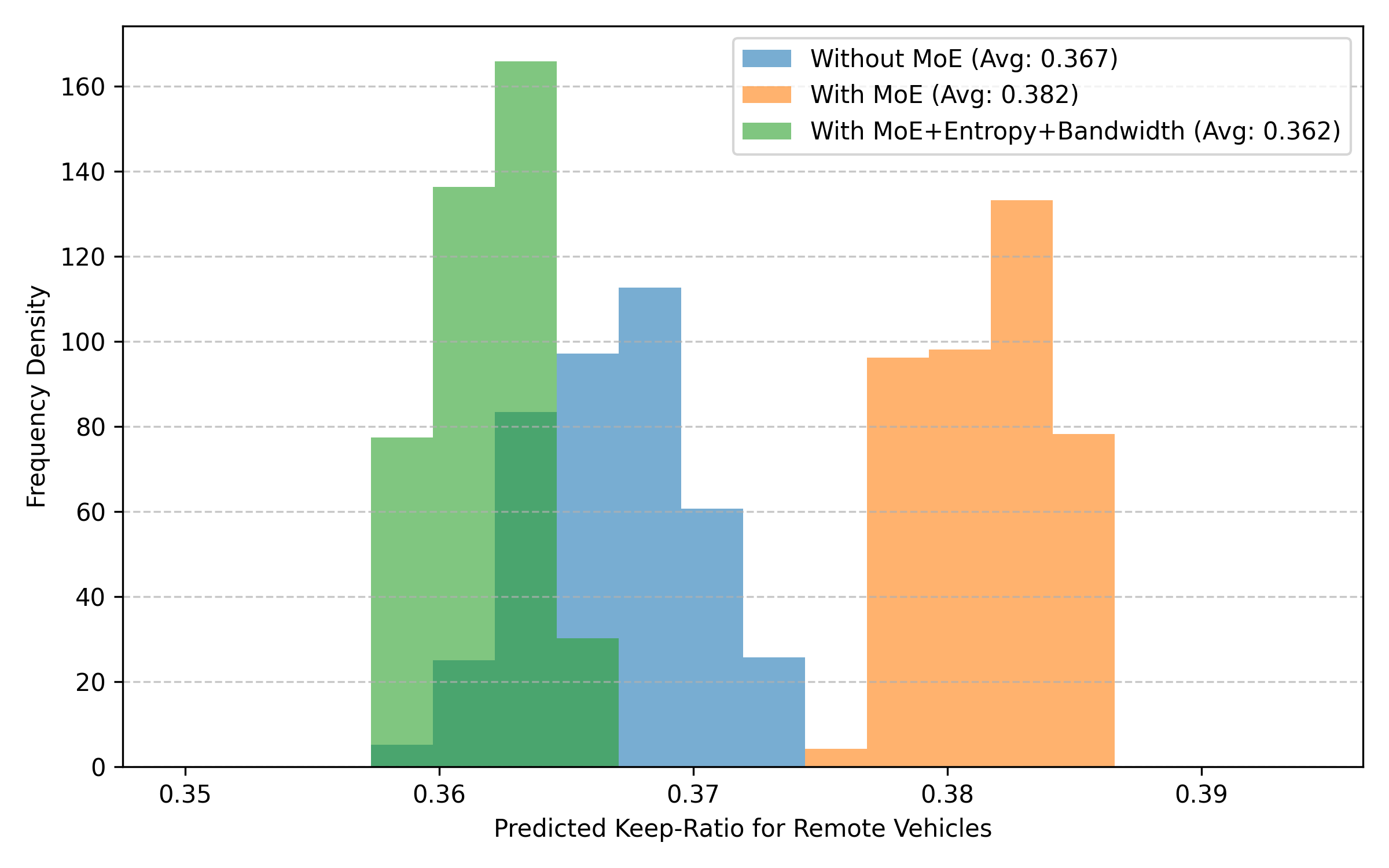}
    \caption{Distribution of predicted keep-ratios for remote
vehicle features by the AGR module under different fu-
sion/training settings. Comparison includes: standard SDPA
fusion trained with detection loss only; MoE fusion trained
with detection loss only; and MoE fusion trained with
additional bandwidth and entropy losses.}
    \label{fig:keep_ratio_histogram_combined}
\end{figure}

Analysis of the keep-ratios from the adaptive grid reduction stage (Figure~\ref{fig:keep_ratio_histogram_combined}) revealed that the MoE baseline variant paradoxically uses a slightly higher average keep-ratio (mean 0.382, orange) than the SDPA baseline (mean 0.367, blue), suggesting less aggressive pruning within the AGR stage itself for this configuration. (The distribution for MoE trained with additional losses (green) is discussed in Sec.~\ref{sec:loss_impact}).
This lower overall bandwidth—despite a higher AGR keep ratio for the MoE baseline—suggests that the primary efficiency gain of the MoE module, when trained only with detection loss, stems from more discriminative upstream feature filtering. We hypothesize that end-to-end training with MoE produces sharper feature or confidence maps, enabling the initial Selective Transmission stage to prune more aggressively and thus reduce the data entering the AGR stage. Consequently, MoE enhances overall communication efficiency through improved upstream filtering in this baseline comparison.

\textbf{Interpretability of MoE fusion:}
Figure~\ref{fig:tsne} presents four complementary views of expert specialization. In the \emph{upper-left} plot, we embed 3\,000 feature tokens (after AGR) into 2-D via t-SNE and colour them by expert gate output, revealing three distinct islands in latent space, each dominated by a different expert. The \emph{upper-right} panel overlays thin grey density contours to emphasize cluster boundaries. In the \emph{lower-left} panel, a pairwise Euclidean-distance heatmap of expert centroids in t-SNE space (0.8–5.7) confirms clear separation. Finally, the \emph{lower-right} bar chart shows the average $\|\mathbf{h}_i\|_2$  of tokens routed to each expert (0.16–0.19), demonstrating that no expert is underutilized. Together, these visuals substantiate that the gating network learns complementary, balanced specializations beyond what AP-vs-bit metrics alone reveal.

\subsection{Impact of Bandwidth and Entropy Loss Terms}
\label{sec:loss_impact} 

While EffiComm with MoE fusion trained solely on detection loss shows improved efficiency (Sec.~\ref{sec:moe_impact}), we further investigated if explicitly optimizing for communication cost and expert diversity during training could yield additional benefits. We experimented with incorporating two additional terms into the training loss function:
$L_{total} = L_{det} + \lambda_{bandwidth} \cdot L_{bw} + \mu_{entropy} \cdot L_{reg}$.
Here, $L_{det}$ is the standard detection loss, $L_{bw}$ is a bandwidth cost proxy (proportion of non-zero features transmitted after reduction), and $L_{reg}$ is an entropy regularization term applied to the MoE gating weights. We used hyperparameters $\lambda_{bandwidth} = 0.05$ and $\mu_{entropy} = 1 \times 10^{-4}$ based on our training configuration. 

Table~\ref{tab:loss_impact} compares the performance of the baseline EffiComm model (MoE fusion, detection loss only) against the variant trained with these additional loss terms ('EffiComm+AL'). Adding the bandwidth and entropy loss terms significantly reduces the average transmitted data per frame from 1.90 MB to \textbf{1.37 MB} and the $\log_2$ communication accordingly, while incurring only a marginal decrease in detection accuracy (0.843 vs 0.840 AP@0.7).

\begin{table}[htbp] 
\footnotesize
\begin{tabular}{l|ccc} 
\toprule
\textbf{Method} & \textbf{AP@0.7} & \textbf{Comm ($\log_2$ count)} & \textbf{Mean BW (MB)}\\
\midrule
EffiComm & \textbf{0.843} & 10.89 & 1.90  \\ 
EffiComm+AL & 0.840 & \textbf{10.41} & \textbf{1.37} \\ 
\bottomrule
\end{tabular}
\caption{Impact of additional loss terms (Bandwidth \& Entropy) on OPV2V test set.}
\label{tab:loss_impact}
\end{table}

\begin{figure}
    \centering
    \includegraphics[width=1\linewidth]{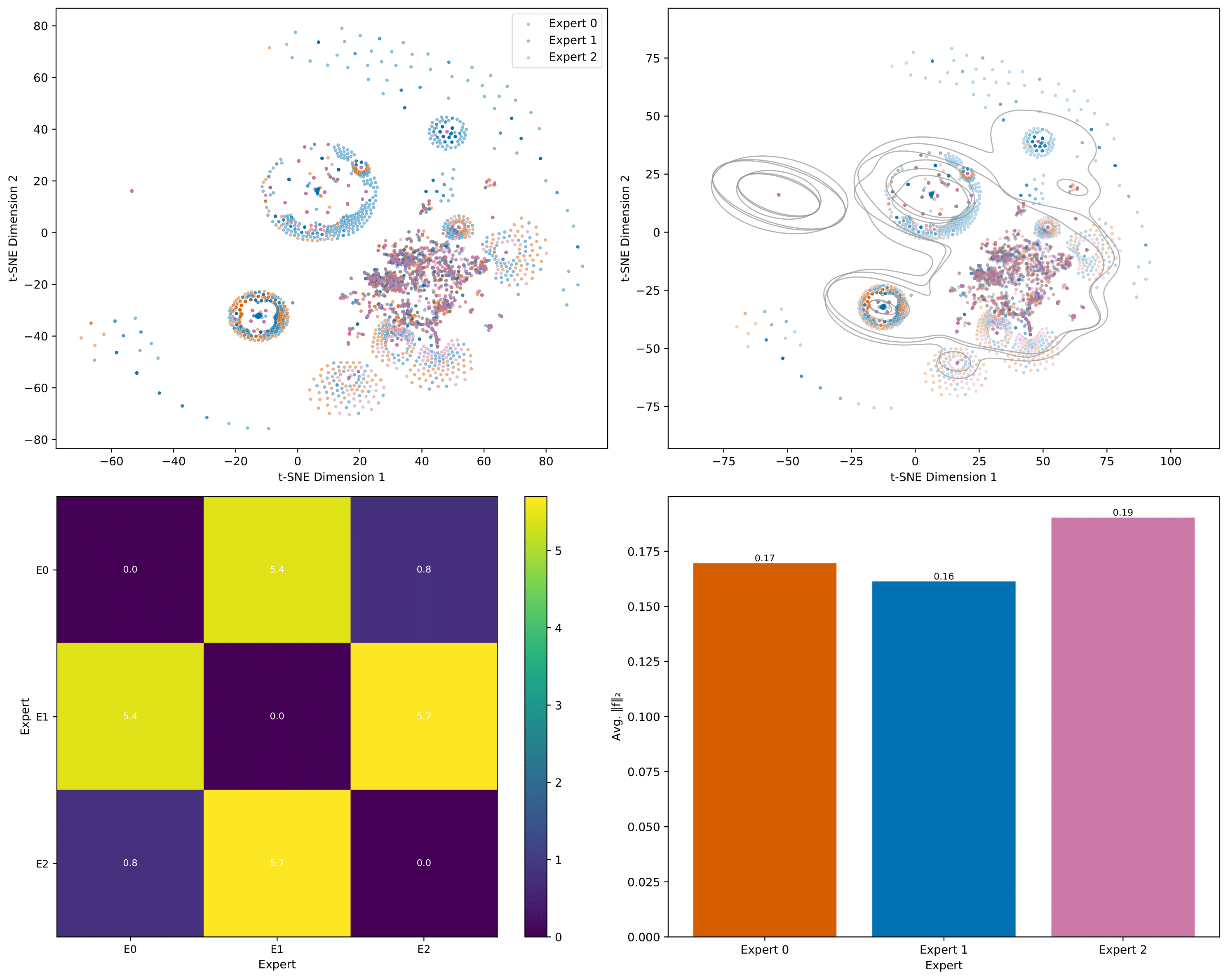}
    \caption{\textbf{Interpretability of MoE fusion.} 
    \emph{Upper-left}: t-SNE of routed tokens coloured by expert ID showing three distinct clusters. 
    \emph{Upper-right}: density contours highlight cluster boundaries. 
    \emph{Lower-left}: centroid-distance heatmap (0.8–5.7) confirms separation. 
    \emph{Lower-right}: average feature norms (0.16–0.19) indicate no expert collapse.}
    \label{fig:tsne}
\end{figure}

\textbf{Keep Ratio Analysis:} We analyzed the effect of these additional loss terms on the keep-ratios predicted by the AGR stage. Referring back to Figure~\ref{fig:keep_ratio_histogram_combined}, the model trained with bandwidth and entropy losses (green distribution) exhibits the \textbf{lowest average keep ratio (0.362)}. This contrasts with the MoE baseline trained only with detection loss (orange distribution, Avg: 0.382) and indicates that the AGR stage becomes more aggressive in its pruning, on average, when explicitly guided by the bandwidth penalty during training. Furthermore, to assess the behavior of the gating network within the MoE module, particularly when trained with the entropy regularization term designed to encourage expert diversity, we examined the average utilization of each expert, as illustrated in Figure~\ref{fig:tsne} \emph{lower-right}.
\section{Future Work}
\label{sec:future_works}
Our work on EffiComm opens several avenues for future research towards more efficient and robust collaborative perception. Refinements to the AGR module could be investigated. The current fully-connected GAT might benefit from alternative graph structures (e.g., distance-weighted or k-NN) or deeper GAT architectures to improve focus and contextual reasoning, especially in dense traffic. Regularizing the MoE router remains an important direction, particularly for larger fleets, to prevent expert collapse and ensure robustness. While our initial investigation using an entropy penalty during training showed combined benefits with a bandwidth penalty (Sec.~\ref{sec:loss_impact}, Table~\ref{tab:loss_impact}), its specific effects warrant further analysis. Further research directions include extending EffiComm to handle heterogeneous sensor data, analyzing performance across more challenging scenarios, optimizing for real-time deployment by reducing latency and overhead, and incorporating multi-object tracking capabilities.
\section{Conclusion}
\label{sec:conclusion}
In this paper, we presented EffiComm, a novel bandwidth-efficient framework for multi-agent collaborative perception. Our approach strategically reduces the transmission load in multi-vehicle communication networks without significantly compromising object detection accuracy. By employing a two-stage reduction pipeline consisting of selective transmission and AGR, along with an attention-based fusion mechanism, EffiComm effectively balances perception performance and communication overhead. To facilitate reproducibility and further research in bandwidth-efficient collaborative perception, the code and trained model weights will be made publicly available on this link \textbf{\underline {https://url.fzi.de/EffiComm}}.
\addtolength{\textheight}{-1.8cm}

\section{Acknowledgment}
\label{sec:acknowledgment}
This paper was created in the “Country 2 City - Bridge” project of the German Center for Future Mobility, which is funded by the German Federal Ministry of Transport.

{\small
\bibliographystyle{IEEEtran}
\bibliography{references}
}

\end{document}